\documentclass[10pt,twocolumn,letterpaper]{article}

\usepackage[table]{xcolor} 
\usepackage{cvpr}
\usepackage{times}
\usepackage{epsfig}
\usepackage{graphicx}
\usepackage{amsmath}
\usepackage{amssymb}

\usepackage{ifthen}
\usepackage{tabularx}
\usepackage{booktabs}
\usepackage{multirow}
\usepackage{siunitx}
\usepackage{caption}
\usepackage{subcaption}  
\newcommand{\PAR}[1]{\vskip4pt \noindent {\bf #1~}}

\newcommand{\nb}[3]{\if\instring{b}{#2}\bfseries\else\fi\if\instring{i}{#2}\itshape{\bfseries}\else\fi\if\instring{h}{#2}\cellcolor{#3}\else\fi#1\if\instring{w}{#2}\phantom{0}\else\fi}
\newcommand{\fakesubfig}[1]{{\kern 0.1em}{\color{red}#1}}

\newcolumntype{C}{>{\centering\arraybackslash}X}

\usepackage[pagebackref=true,breaklinks=true,letterpaper=true,colorlinks,bookmarks=false]{hyperref}

\cvprfinalcopy 

\def\cvprPaperID{6} 

\ifcvprfinal\fi
\begin{document}

\title{Towards a Principled Integration of Multi-Camera Re-Identification and Tracking through Optimal Bayes Filters}

\author{Lucas Beyer\thanks{Equal contribution. This work was funded by the ERC Starting Grant project CV-SUPER (ERC-2012-StG-307432) and the EU project STRANDS (ICT-2011-600623).},\hspace*{2pt} Stefan Breuers\footnotemark[1]\hspace*{4pt}, Vitaly Kurin, and Bastian Leibe\\
Visual Computing Institute\\
RWTH Aachen University\\
{\tt\small last@vision.rwth-aachen.de}, {\tt\small vitaliykurin@gmail.com}
}

\maketitle

\begin{abstract}
    With the rise of end-to-end learning through deep learning, person detectors and re-identification (ReID) models have recently become very strong.
    Multi-target multi-camera (MTMC) tracking has not fully gone through this transformation yet.
    We intend to take another step in this direction by presenting a theoretically principled way of integrating ReID with tracking formulated as an optimal Bayes filter.
    This conveniently side-steps the need for data-association and opens up a direct path from full images to the core of the tracker.
    While the results are still sub-par, we believe that this new, tight integration opens many interesting research opportunities and leads the way towards full end-to-end tracking from raw pixels.
    Code and models for all experiments are publicly available\footnote{https://github.com/VisualComputingInstitute/towards-reid-tracking}.

\end{abstract}


\section{Introduction}\label{sec:intro}
\begin{figure}
       \centering
       \begin{subfigure}[b]{\linewidth}
               \centering
               \includegraphics[height=4cm]{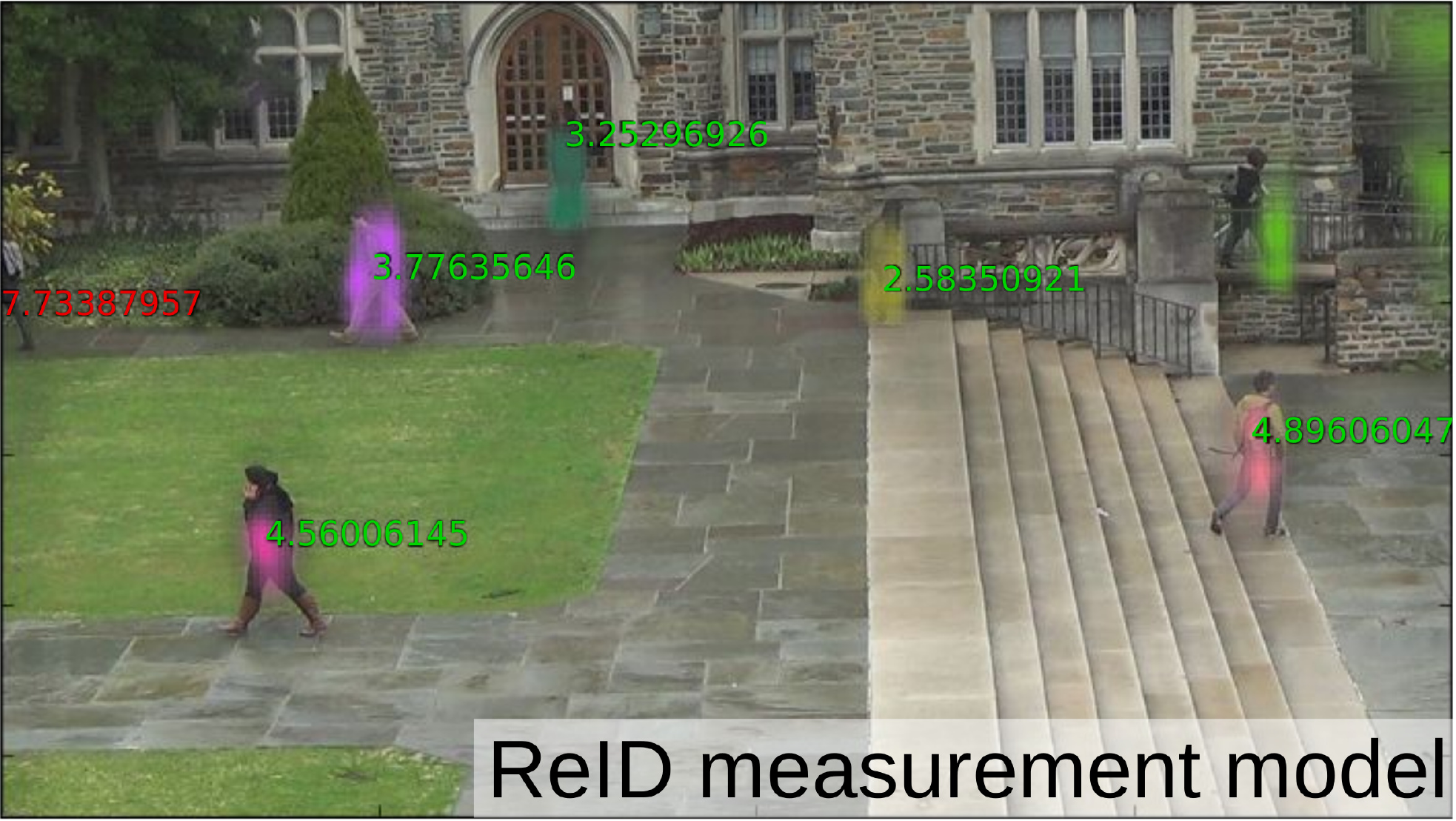}
               \caption{Person-specific observations from the ReID model.}
               \label{fig:covergirl_reID}
       \end{subfigure}\\
       \begin{subfigure}[b]{\linewidth}
               \centering
               \includegraphics[height=4cm]{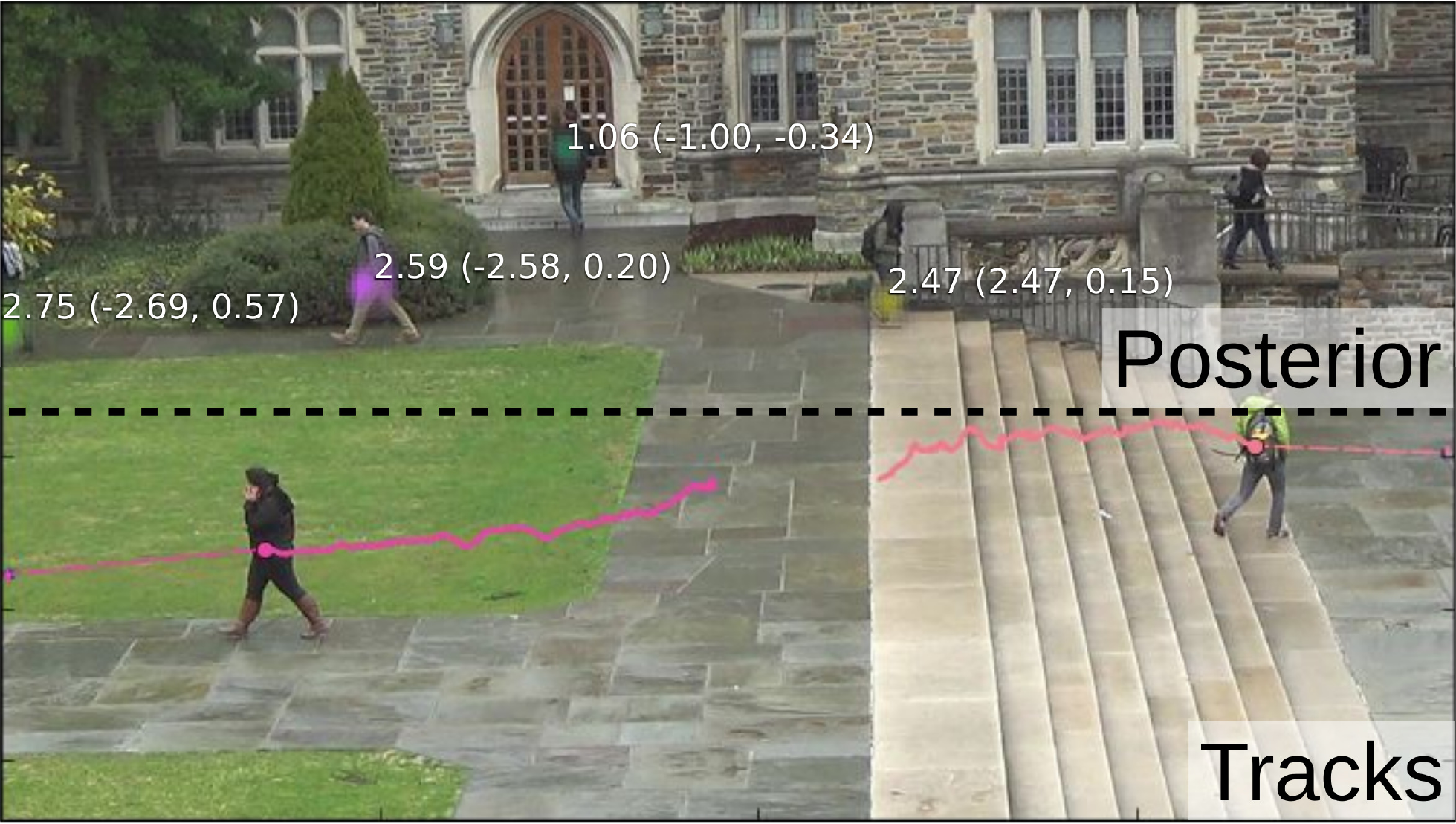}
               \caption{Top: posterior location probabilities, bottom: output.}
               \label{fig:covergirl_postAndTracks}
       \end{subfigure}
       \caption{Our formulation uses a) ID-specific measurements to b) propagate the full posterior probability of person locations, which is used to produce tracks.
                The red and green numbers are distances in ReID embedding-space (lower corresponds to a better match), white numbers are velocities in $\left[px/fr\right]$.
                Best viewed in color.}
       \label{fig:covergirl}
\end{figure}

    In the past years, most subfields in computer vision have benefited immensely from the move towards end-to-end learning from raw pixels through the deep learning framework.
    Tracking should be no exception to this, but several obstacles make this change non-trivial:
    \textit{i)} the discrete boundary between the image and tracks which is formed by detections,
    \textit{ii)} the data-association problem, \ie, which measurements belong to which target, and
    \textit{iii)} track management, meaning that the tracker's output is an unordered set whose size changes.
    The latter two have recently been adressed in an elegant way by Milan~\etal~\cite{milan2016online} using recurrent neural networks both for learning to solve the data-association problem as well as for track management.
    Unfortunately, they are using detections as a starting point and thus, while moving a big step forward, they do not perform full end-to-end ``pixels to tracks'' learning yet.

    Since person ReID has recently become very strong~\cite{HermansBeyer17Arxiv,Geng16Arxiv,ZhengZ16Arxiv}, we want to explore the possibilities of integrating ReID and tracking.
    Specifically, in this work we propose a principled way of removing the first boundary mentioned above, namely that formed by detections, by integrating a strong ReID model into the optimal Bayes filter formulation of tracking.
    As an added benefit, this formulation also obviates the need for a data-associaton step altogether.
    Because, to the best of our knowledge, the formulation is new, it comes with a lot of unsolved questions and opens up many new research opportunities.
    We see this work as orthogonal to that of Milan~\etal~\cite{milan2016online} and look forward to integrations of the two ideas leading to full end-to-end tracking.

\section{Related Work}\label{sec:related}
Both ReID~\cite{vezzani2013people,ZhengL16Arxiv} and tracking~\cite{LuoZhaoKimXiv14} come with a vast amount of previous work, for which we refer the interested reader to the cited surveys for a full overview.
We will focus on systems that utilize and combine deep learning for both tracking and ReID.

Early work on using the person identity to help tracking is presented by~\cite{track_pirmpt}, where classical color-based methods are used.
More recent works use CNNs for appearance-based tracking~\cite{chen2016cnntracker,li2016deeptrack,wang2015transferring}, but typically come with the need of fine-tuning person-specific models online, which might be effective but very costly and is subject to model drift.
Leal-Taix\'{e}~\etal present a siamese network to link similar person boxes~\cite{leal2016learning}, and by this address the complex problem of data association, albeit with the dependency on a detector.
Alahi~\etal~\cite{alahi2016social} focus on deep prediction models for multi-target tracking with social long-short term memories (LSTMs).
The work of Sadeghian~\etal~\cite{sadeghian2017tracking} goes a step further by training LSTMs for ReID, motion, and interaction of persons, but still operates on discrete detector bounding boxes.

One of the conceptual differences between the aforementioned works and our explorative work is exactly this dependency on a person detector, providing discrete boxes as starting points.
This gives limited state information regarding position, and makes tracking an instance of the complex data association problem.
First work towards end-to-end tracking by learning this data association was recently done by~\cite{milan2016online}.
But as mentioned in the introduction, we want to drop both the data association and discrete box representations, and instead keep track of the full belief for each person by leveraging recent ReID models.

\section{Method}\label{sec:method}

First, we briefly summarize a recent, very strong person-ReID model.
We then review the optimal Bayes filter forming the theoretical background in tracking and present some basic approaches of data association which are usually utilized for detection-based multi-object tracking.
After briefly showing the naive way in which a Re-ID model is used as an appearance model to help data association, we take a step back and derive a more principled combination.
Here, we again follow the bottom-up formulation of tracking as an optimal Bayes filter.
We argue towards being independent of a detector, by using a strong ReID model for person-specific measurements.
This both avoids the problem of data association and a discrete state representation in form of bounding boxes, by combining full probability maps, without assuming any family of density functions.

\subsection{Tracking Formulation}\label{sec:tracking_formulation}
In order to work towards a principled integration of MTMC and ReID, we need to revisit the origin of current tracking formulations, namely the optimal Bayes filter.
At its core, the optimal Bayes filter consists of the application of the Bayes rule to time-series:

\newcommand{\given}{\,\mid\,}

\begin{equation}\label{eq:bayesfilt}
    P(X_{t}{\given}z_t, z_{1:t-1}) \propto
        \overbrace{P(z_{t}{\given}X_{t}, z_{1:t-1})}^{\textnormal{new measurement}}
        \overbrace{P(X_{t}{\given}z_{1:t-1})}^{\textnormal{belief propagation}}.
\end{equation}

The resulting equation can be interpreted as propagating the current belief about the state forward (\emph{predict} step) and updating it with new measurements as they come (\emph{update} step).
Uppercase letters represent random variables, while lowercase letters represent concrete values, or instantiations.



\PAR{Predict step.}
The belief-propagation term can be further decomposed using the law of total probability, and the Markov assumption that $P(X_t{\given}x_{t-1}, z_{1:t-1}) = P(X_t{\given}x_{t-1})$, \ie that the previous state contains all information about all past measurements:

\begin{equation}\label{eq:track_predict}
    P(X_{t}{\given}z_{1:t-1}) =
        \int{\overbrace{P(X_{t}{\given}x_{t-1})}^{\textnormal{dynamics model}}
             P(x_{t-1}\mid z_{1:t-1})} dx_{t-1}.
\end{equation}

The above equation computes a belief about the current state $X_t$ based solely on the previous state $X_{t-1}$ and all past measurements $z_{1:t-1}$.
The term $P(X_{t}{\given}x_{t-1})$ represents the dynamic model and is typically computed via a motion model in MTMC, \eg constant velocity for pedestrians.

\PAR{Update step.}
Now the propagated belief is updated with a new incoming measurement $z_t$. Here, Bayes rule is applied to give the new posterior state $P(X_{t}{\given}z_{1:t})$:
\begin{equation}\label{eq:track_update}
    P(X_{t}{\given}z_{1:t}) =
        \frac{\overbrace{P(z_{t}{\given}X_{t}, z_{1:t-1})}^{\textnormal{measurement model}}P(X_{t}{\given}z_{1:t-1})}
             {\int{P(z_{t}{\given}x_{t}, z_{1:t-1}) P(x_{t}{\given}z_{1:t-1}) dx_{t}}}
\end{equation}
This corresponds to Eq.~\ref{eq:bayesfilt}.
Note that if no measurement is available, $P(z_t{\given}X_t, z_{1:t-1})$ is uniform and thus cancels out after normalization, meaning that the prediction is used as posterior: $P(X_t{\given}z_{1:t}) = P(X_t{\given}z_{1:t-1})$.

In tracking, the state usually contains position, velocity, and appearance and the measurements, which consist of detector bounding-boxes, provide positional information.
The multi-target tracking problem can be formulated in two ways: either as a single Bayes filter, whose state $X$ and the measurements $z$ contain all known persons' states and all available measurements, in which case the dynamics model can account for social interactions and the measurement model needs to associate measurements to persons;
or it can be formulated on a per-track basis, where one Bayes filter is run per person and its state $X$ contains only that person's state, hence the association of individual measurements $z$ to tracks is done ``from the outside'', and considered given to the Bayes filter.

It is exactly this \emph{data association} (DA) that makes up the core of many tracking methods, be it ``from the outside'' or as part of the measurement model $P(X_t{\given}z_{1:t-1})$.
Given a set of measurements $\mathcal{Z}_{t}$ at time $t$ we do not know, which measurement $z_{t} \in \mathcal{Z}_{t}$ should contribute to which of the current states.

A common, relatively straightforward way of performing DA is by introducing another Markov assumption in the measurement model: given the current state, past measurements give no additional information about the current measurement: $P(z_t{\given}X_t, z_{1:t-1}) = P(z_t{\given}X_t)$.
Under this assumption, the optimal assignment of detections to their nearest tracks can be computed relatively efficiently at each individual time-step $t$ using the Hungarian algorithm~\cite{munkres1957algorithms}.
Joint Probabilistic DA (JPDA~\cite{fortmann1980multi,Rezatofighi_2015_ICCV}) relaxes the Hungarian algorithm's hard assignments and instead considers a soft, probabilistic assignment of measurements to tracks.
More sophisticated DA models such as Multiple Hypothesis Tracking (MHT~\cite{reid1979algorithm,Kim_2015_ICCV}) and MCMCDA~\cite{OhMCMCDA} get rid of the aforementioned Markov assumption and consider the evolution of associations over time. 

A second important consideration when implementing a tracking system as a Bayes filter is the choice of representation for the state.
The \emph{Kalman filter}~\cite{kalman1960new,welch2006introduction} is probably the most widely used implementation of a Bayes filter, it assumes that the beliefs follow a Gaussian distribution and can thus be parametrized by their means and covariances.
Furthermore, the Kalman filter requires both the dynamic model and the measurement model to be linear functions with additive Gaussian noise.
Under these very strong and restrictive assumptions, the Kalman filter allows for an efficient and compact implementation of a tracking system.
Alternative state representations are, amongst others, particles and histograms, leading to \emph{particle filters}~\cite{isard1998condensation} and \emph{histogram filters}~\cite{macdonald1997hidden}, respectively.
While these representations do not allow analytical solutions and thus result in less compact and efficient implementations, they are by far more expressive and can model full, diverse posterior probabilities.


In the end, to implement a Bayes filter, we need an initial belief $P(X_0)$, a belief propagation (dynamics) model $P(X_{t}{\given}x_{t-1})$ and a measurement model $P(z_{t}{\given}X_{t}, z_{1:t-1})$.
We will discuss all these in light of integrating MTMC and ReID in the following.

\subsection{ReID as Appearance Model}\label{sec:method_naive}

Person Re-Identification (ReID) consists of matching images of the same person taken under different conditions, such as different cameras, lighting, and pose.
It is often formulated as a retrieval task: given one image of a person, sort all other images (of both the same and other persons) and retrieve the best matches.
As such, datasets used for tracking across multiple cameras can also be used for learning and evaluating ReID systems.

Recently, the field of person ReID has gone through the same deep-learning revolution so many other fields have gone.
Several very strong models have been proposed~\cite{HermansBeyer17Arxiv,Geng16Arxiv,ZhengZ16Arxiv}, all of which profit from the recent introduction of large-scale person ReID datasets, namely Market-1501~\cite{ZhengL15ICCV} and MARS~\cite{ZhengL16ECCV}.
The introduction of large-scale multi-camera tracking datasets such as DukeMTMC~\cite{ristaniECCV16} thus naturally suggests learning such ReID models on tracking data and using them as part of a tracker's appearance model.

Given a ReID model that scores the similarity of two person crops, an immediate way to include it in a tracking framework is as additional information to the data association.
Regardless of the DA method used, whenever the metric distance between tracks and detections is typically used, this distance can be replaced by, or augmented with, the ``appearance distance'' provided by the ReID model.~\cite{Kim_2015_ICCV,yang2012online,leal2016learning}
For instance, one of many ways of combining these is by computing a combined distance
\begin{equation}
    d(I_1, I_2) = \frac{d_\text{pos}(I_1, I_2)}{N_\text{pos}} \frac{d_\text{app}(I_1, I_2)}{N_\text{app}},
    \label{eq:comb_dist}
\end{equation}
where $N_\text{pos}$ and $N_\text{app}$ bring the two distances into a common scale.
This combined distance can then be used in place of $d_\text{pos}$ alone in the DA method of choice.

\subsection{Principled Integration of ReID and Tracking}\label{sec:method_integration}

\begin{figure*}
	\centering
	\includegraphics[height=10cm]{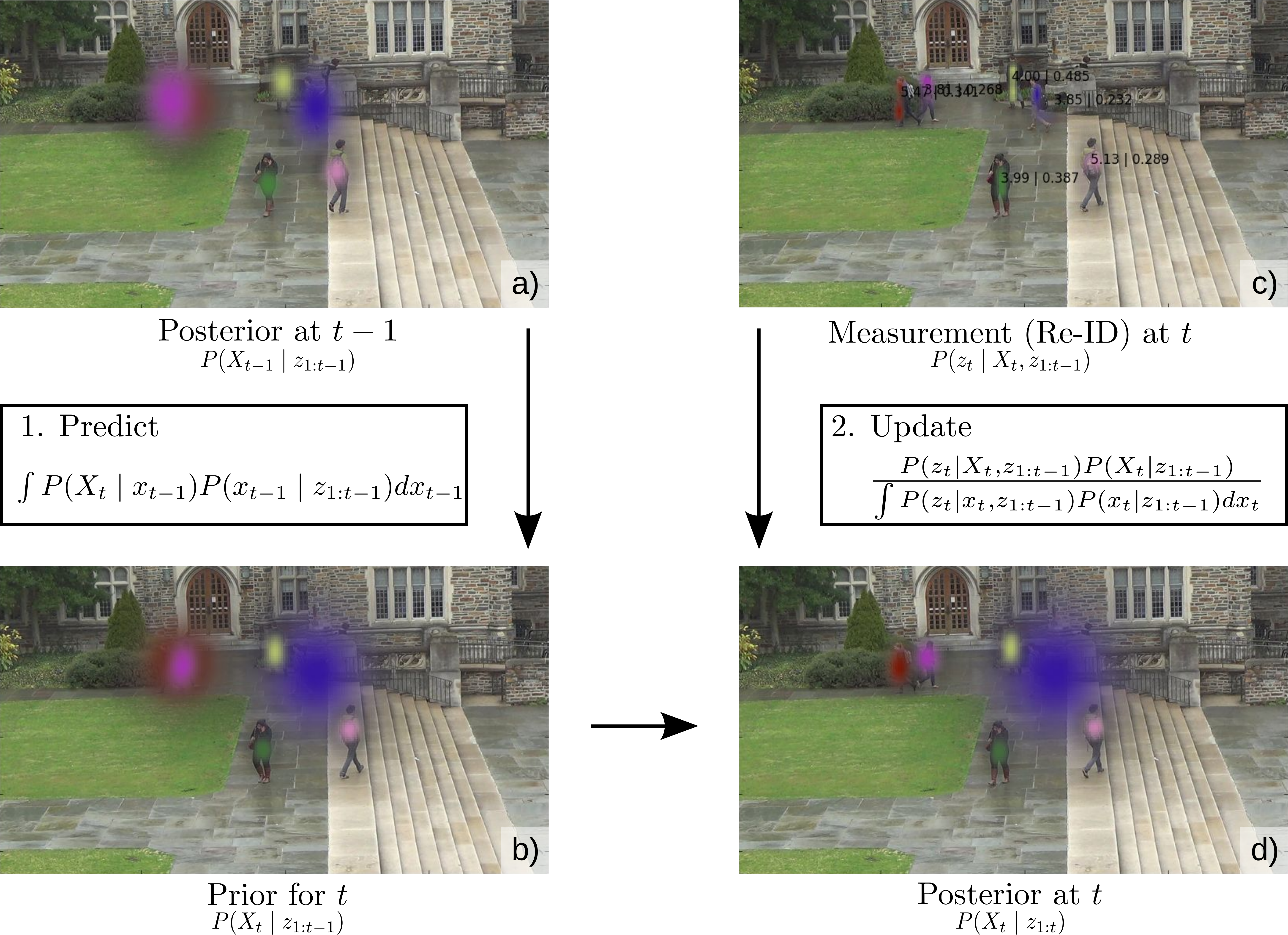}
    \caption{Full overview illustrating our pricipled integration of ReID inside an optimal Bayes filter for tracking.}
    \label{fig:overview}
\end{figure*}

The above way of integrating ReID and MTMC-Tracking appears rather ad-hoc, though, and we aim to explore a more principled integration approach.
An overview of the proposed approach is shown in Fig.~\ref{fig:overview}.
We propose to represent beliefs and measurements about spatial location as spatial histograms (or \emph{probability maps}) and thus implement the location's Bayes filter as a histogram filter.
This means that expressions such as $p(X_t)$ are represented by a discrete table of probabilities as opposed to, for example, a mean and a covariance.
We do, however, represent beliefs and measurements of the velocity as Gaussian distributions parametrized by their mean and covariance, as is common in the literature.
The reason for these choices will become apparent in the following.

\PAR{Measurement model.}
In~\cite{HermansBeyer17Arxiv}, Hermans and Beyer present a ReID model $f_\theta$ which maps an image of a person $I_p$ into an \emph{embedding vector} $e_p = f_\theta(I_p)$.
Practically speaking, the model learns to represent people by $128$-dimensional \emph{embedding vectors} such that the embeddings of different pictures of the same person are nearby in the Euclidean sense, whereas embeddings from different persons are far apart.
The function $f_\theta$ is represented by a deep convolutional neural network (CNN) and optimized using a variant of the triplet loss.

By converting all fully-connected layers of the CNN to $1\times1$ convolutions, the network can be efficiently applied on a full image $I$ in a convolutional way, resulting in a low-resolution \emph{embedding map} $\mathbf{E}_I = e_{i,j}$ of the image.
Each location in this embedding map is a $128$-dimensional vector $e_{i,j}$ representing the identity of the correspondingly centered crop in the input image.
After embedding the full image once, it is cheap to compute an individual measurement for each track by computing the distance between the track's embedding $e_p$ and the image's embeddings $\mathbf{E}_I$, resulting in a person-specific embedding distance map $\mathbf{D}_I(e_p) = \left(\|e_{i,j} - e_p\|\right)_{i,j}$.
A low distance in $\mathbf{D}$ signifies the presence of a similarly-looking person, whereas a high distance does not, cf.~Fig.~\ref{fig:covergirl_reID}.

This person-specific distance map can be used as measurement in the Bayesian filter.
Specifically, we define the measurement model for a single track as
\begin{equation}
    P(z_t{\given}X_t, z_{1:t-1}) = \operatorname{softmin}(\mathbf{D}_I(f(X_t, z_{1:t-1}))),
\end{equation}
where the track's embedding $e_p = f(X_t, z_{1:t-1})$ can be kept up-to-date in many different ways, but for simplicity, we use $e_p = f_\theta(z_1)$ in this work.
Figures~\ref{fig:covergirl_reID} and~\ref{fig:overview}\fakesubfig{c} show multiple such person-specific measurement maps overlaid, in a different color for each person.

Note how this way of obtaining an individual measurement for each individual person completely sidesteps the need for data association.

\PAR{Predict.}
Now that we have a measurement model, the next necessary component for a full Bayesian filter is the \emph{predict} step, which uses the dynamics model in order to make a prediction of the new state (see Eq.~\ref{eq:track_predict}) before new measurements come in.
In a typical Kalman filter-based tracker, this is done by moving the belief about the position using the belief about the velocity and increasing the uncertainty.
Remember that we parametrize the velocity as a Gaussian using a mean and a covariance.
In this case, we can exactly implement the \emph{predict} step in Eq.~\ref{eq:track_predict} by convolving the positional belief-map (posterior from the previous time step) with a Gaussian kernel whose mean and covariance correspond to those of the current belief about the velocity.
This step is the operation leading from Fig.~\ref{fig:overview}\fakesubfig{a} to Fig.~\ref{fig:overview}\fakesubfig{b}.

\PAR{Combination.}
Now that we have both the new position measurement and the position belief propagation as probability maps, Eq.~\ref{eq:track_update} tells us that all we need to do is multiply them element-wise and re-normalize the resulting to get the posterior probability-map of the position.
This completes the Bayes filter for the position.

\PAR{Output.}
Since most applications and evaluation frameworks require discrete locations as outputs of a tracker, as opposed to full posterior-maps, a last step is necessary.
The posterior can be turned into a prediction by taking its peak, which would correspond to performing MAP prediction.
Alternatively, the expected location can be computed simply by taking the expectation of the location under the posterior.
More interestingly, various higher-level components could potentially make use of the full posterior in order to compute expectations, as opposed to relying on point-estimates.

\PAR{Velocity.}
We have so far mostly ignored the velocity, except for stating that in this initial work, we propose to represent it as a Gaussian probability and update it following classic Kalman filter approaches.
The only thing that changes is the way in which a velocity measurement is retrieved.
Measuring the velocity is one specific example of state building on top of the position, and it can now be estimated in various ways using the position's posterior maps at $t-1$ and $t$.
A point-estimate can be computed by comparing their MAPs, \ie, their peaks.
An expectation of the velocity can be computed by using the full posterior, or a full velocity-map could be estimated through optical flow.
Since it is not a focus of this work, we go for the simplest option of a point-estimate and leave the other possibilites as future work.

\PAR{Recap.}
To summarize, we propose the following approach to integrate ReID and tracking in a principled way:

\begin{itemize}
\item Use the ReID model to generate track-specific observations and thus sidestep the need for data association.
\item Represent the position using probability maps, which combine well with the ReID's observations, and lead to a histogram filter for the position.
\item Adapt the predict and update steps to work with probability maps.
\item Represent the velocity as Gaussian parametrized by a mean and a variance, as is commonly done.
\end{itemize}

This leads to an elegant formulation, and efficient implementation with rich state representations.

\section{Implementation and Evaluation}\label{sec:impl}
Many datasets are available to evaluate tracking, such as the MOTChallenge benchmarks~\cite{MOTChallenge2015,MOTChallenge2016}, KITTI~\cite{Geiger2012CVPR} for urban street scenes, and PETS~\cite{data_pets} for surveillance scenarios.
We choose the DukeMTMC dataset~\cite{ristaniECCV16,ristaniACCV14,soleraCSVT16} for our investigation as it is the only one that is large enough to be suitable for learning a deep ReID model.
It contains almost $3\,000$ person identities across 8 static cameras, recording 1:25 hours of video at 60 FPS each.
It also comes with DPM~\cite{felzenszwalb2008discriminatively} detection boxes and defines clear-cut train-test splits.
As we need ground-truth boxes for some of our experiments, we perform our evaluation on the predefined \emph{trainval-mini} section of the training split.
Importantly, we exclude all persons which appear in that validation set from training, such that the ReID network as never before seen the persons presented to it during tracking.


For a quantitative evaluation, we follow the dataset's recommendation and report the widely accepted CLEAR metrics~\cite{BS2008}: MOT accuracy (MOTA) and precision (MOTP), which depend on the number of false positives (FP), false negatives (FN), and identity switches (IDS).
Furthermore, the number of mostly tracked (MT, $>80\%$ of time) and mostly lost (ML, $<20\%$ of time) targets is reported as suggested by~\cite{mtptml}, as well as ID-specific scores (recall IDR, precision IDP, F1-score IDF1) used by the DukeMTMC dataset~\cite{ristaniECCV16}.


Table~\ref{tab:all_results} summarizes all main experiments, which we will discuss step-by-step in the following, alongside with further implementation details.

\subsection{Regressed Bounding-box Output}\label{sec:impl_regression}
Almost all tracking-by-detection algorithms operate on bounding boxes~\cite{breuers2016exploring}.
While we argue for getting rid of this discrete state representation as the barrier between the image and tracks, it can still be necessary for a tracker to \emph{output} bounding boxes depending on the application and follow-up tasks.
For example, a subsequent system could estimate the location of a person's head using the bounding-box and subsequently infer the person's head orientation~\cite{beyer2015biternion,benfold2011unsupervised}.
On the other side, \eg, for the task of robotic navigation~\cite{triebel2016spencer,hawes2016strands}, a foot point on the ground plane is enough, and bounding-boxes are completely unnecessary.

\begin{figure}
	\centering
	\begin{subfigure}[b]{0.24\textwidth}
		\centering
		\includegraphics[height=2.5cm]{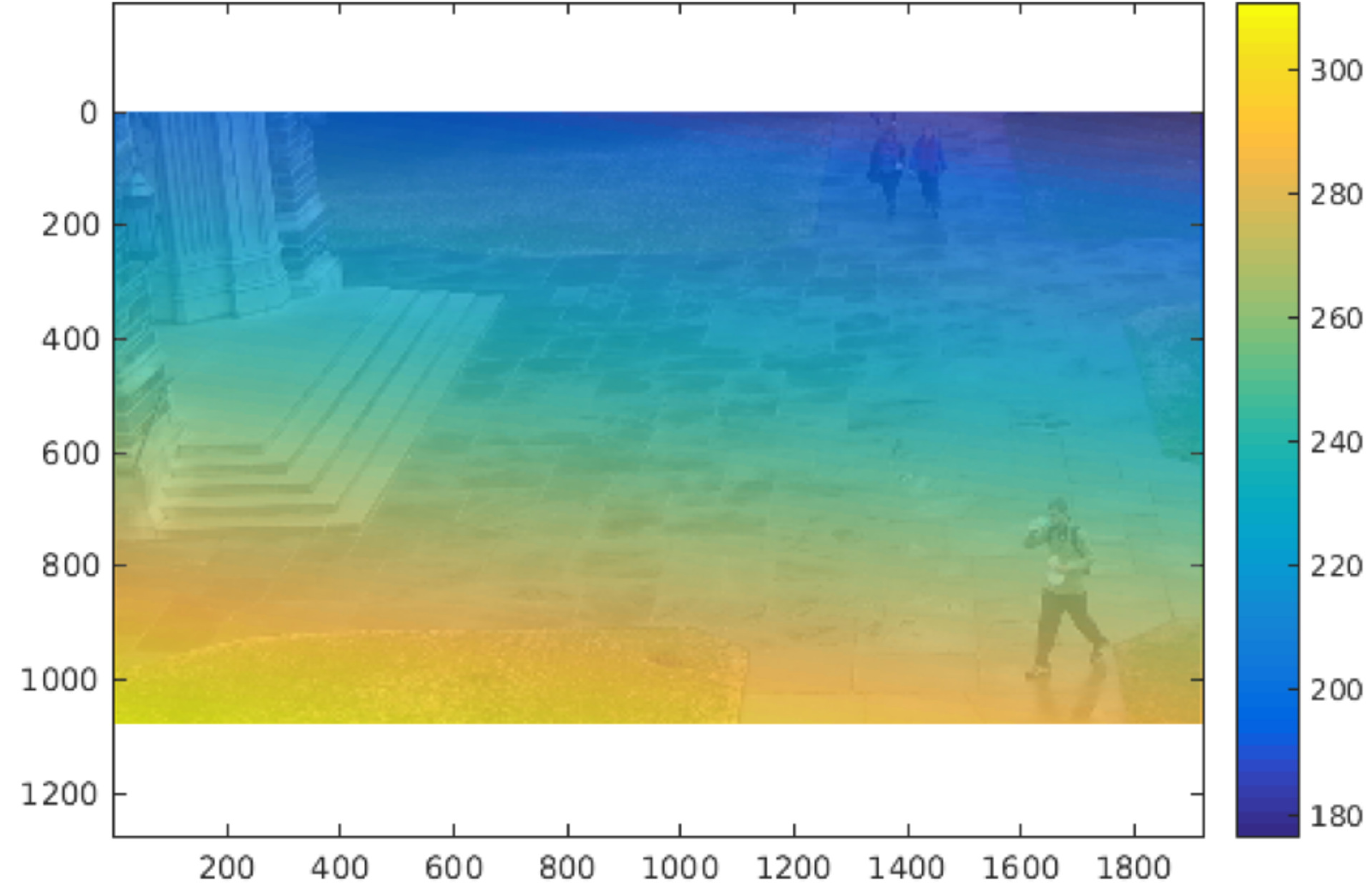}
		\caption{}
		\label{fig:bbreg_example}
	\end{subfigure}%
	\begin{subfigure}[b]{0.24\textwidth}
		\centering
		\includegraphics[height=2.5cm]{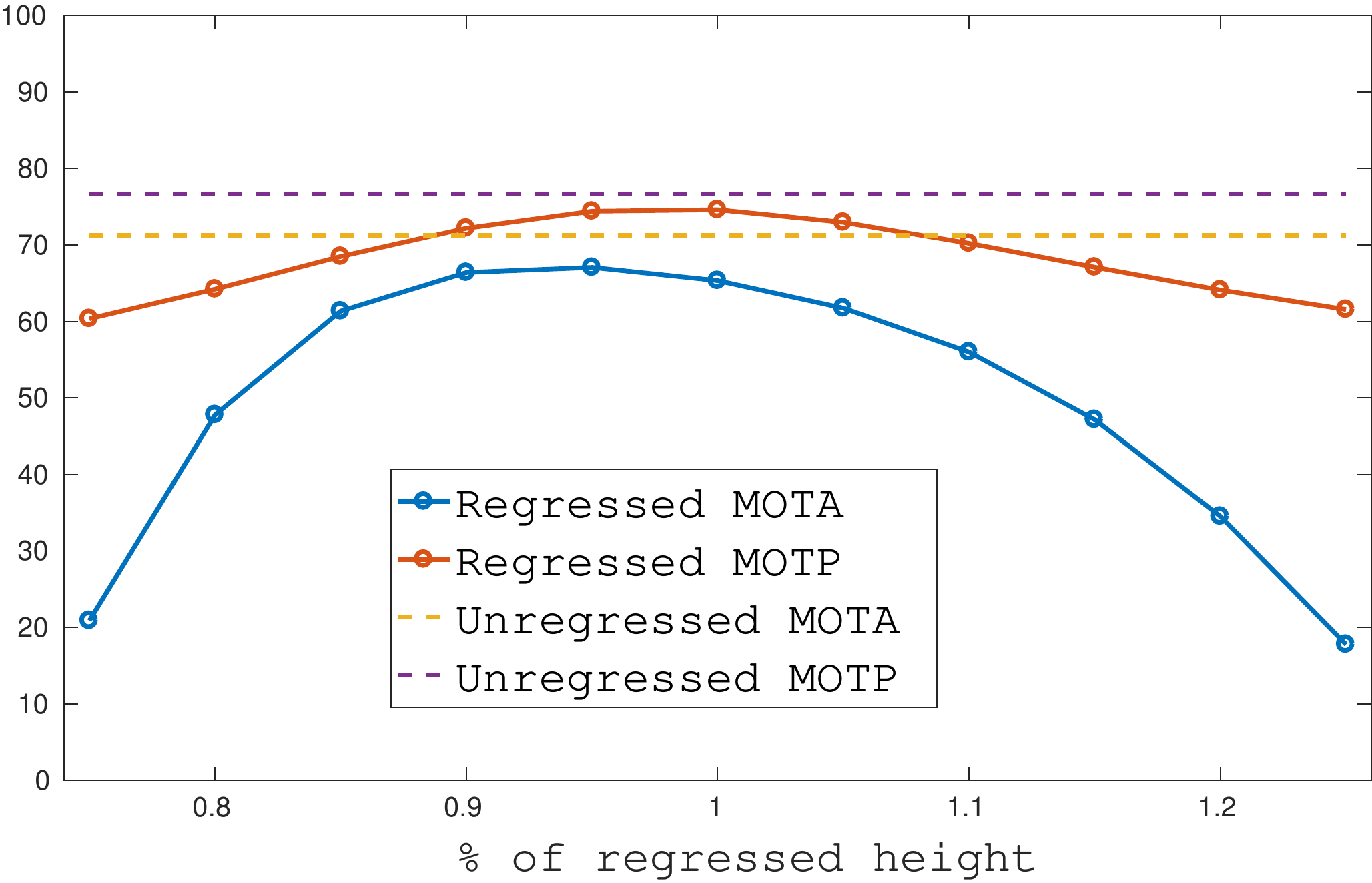}
		\caption{}
		\label{fig:bbreg_plot}
	\end{subfigure}
	\caption{The camera-specific bounding-box regression. a) Bounding box height as a function of center-point location. b) Development of scores when scaling our regression.}
	\label{fig:bbreg}
\end{figure}

Since we evaluate our approaches in image coordinates on the DukeMTMC dataset, outputting bounding-boxes is mandatory following the CLEAR MOT metrics~\cite{BS2008} and, as shown by~\cite{MilanSchindlerRothCVPR13}, it is important to keep evaluation consistent.
Thus, as our system gives full probability maps, mostly centered on the middle point of a person, we learn a camera-specific bounding box regressor for each camera (cf.~Fig.~\ref{fig:bbreg_example}).
This effectively reports the same average person height for all persons, depending on their position in the camera image, while the width is defined through the average aspect ratio and set to $0.4$ times the height.
This is obviously suboptimal and ignores some volumetric information that is, \eg, given by a detector.
But as we actually want to drop the dependency on these, good bounding boxes shall not be the focus of this exploration either.

Still, we perform a quantitative evaluation of the effect of this decision by comparing the performance of the dataset's baseline BIPCC~\cite{ristaniECCV16}, a batch tracking method based on correlation clustering, with its performance using the regressed bounding boxes (cf.~Tab.~\ref{tab:all_results}).
Unsurprisingly, all scores worsen, although only the number of FP does dramatically so.
Overall, the general MOTA score decreases by 10 percentage points, but the good news is that MOTP increases by 3 percentage points.
This means, even if we lose some correct boxes, the average Interesection over Union (IoU) of the remaining matches is still fine.
As a precise localization, \ie, a good MOTP score, is the main aspect of bounding boxes as output, we go with the unscaled regression even though a scaled height gives slightly better MOTA scores (cf.~Fig.~\ref{fig:bbreg_plot}).


	      \begin{table*}[t]
            \footnotesize
            \setlength{\tabcolsep}{1pt}
            \setlength{\extrarowheight}{5pt}
            \renewcommand{\arraystretch}{0.75}
            \centering
            \begin{tabularx}{\linewidth}{p{1.6cm} Cp{2pt}Cp{2pt}C p{25pt} Cp{2pt}Cp{2pt}C p{25pt} Cp{2pt}Cp{2pt}C p{25pt} Cp{2pt}C}
                \toprule[1pt]
                          & IDF1 && IDP  && IDR  && total && MT && ML && FP && FN && IDS &&  MOTA && MOTP\\
                \midrule[0.5pt]
                BIPCC \cite{ristaniECCV16} & 75.0 && 85.5 && 66.8 && 1489 && 912 && 96 && 40726 && 301360 && 378 &&  76.7 && 71.3 \\
                + regression               & 72.5 && 81.3 && 65.5 && 1489 && 869 && 94 && 90458 && 322373 && 396 &&  65.3 && 74.6 \\
                \arrayrulecolor{lightgray}\midrule[0.25pt]\arrayrulecolor{black}
                NN-KF     & 28.5 && 35.5 && 23.8 && 1489 && 334 && 238 && 176293 && 568559 && 10658 &&  36.6 && 73.6 \\
                + GT init & 57.9 && 59.1 && 56.9 && 1489 && 624 && 119 && 369121 && 413904 &&  1603 &&  34.2 && 71.8 \\
                + ReID    & 66.9 && 75.7 && 59.9 && 1489 && 648 && 150 && 213544 && 462723 &&   799 &&  43.2 && 71.7 \\
                only ReID & 60.6 && 64.9 && 56.8 && 1489 && 586 && 166 && 344224 && 493665 &&  1332 &&  29.6 && 71.7 \\
                \arrayrulecolor{lightgray}\midrule[0.25pt]\arrayrulecolor{black}
                Full      & 42.2 && 42.0 && 42.4 && 1489 && 389 && 347 && 678844 && 666813 &&  2137 && -13.0 && 67.6 \\
                + entropy & 40.1 && 63.2 && 29.4 && 1489 && 256 && 696 && 192113 && 830097 &&   883 &&  14.2 && 68.4 \\
                \bottomrule[1pt]
            \end{tabularx}
            \caption{Comparison of all considered systems (scores averaged over all single camera results).}
            \label{tab:all_results}
            \vspace*{-15pt}
        \end{table*}

\subsection{Baseline NN-tracker}\label{sec:impl_baseline}
As next step, we present a baseline for tracking as Bayes filter via the classical DA approach.
As mentioned in Section~\ref{sec:tracking_formulation}, there are many sophisticated DA methods~\cite{fortmann1980multi,Rezatofighi_2015_ICCV,reid1979algorithm,Kim_2015_ICCV,OhMCMCDA}.
Since we ultimately will get rid of the DA step by using ReID, we keep this part relatively simple by using a nearest-neighbor Kalman Filter (NN-KF).
While obviously not a state-of-the-art method, a well-designed NN-KF method has been shown to perform competitively in real-world scenarios~\cite{lindermulti}.

For comparability, our NN-KF's input is the center point of the detections, and the output bounding boxes are regressed as mentioned in Section.~\ref{sec:impl_regression}.
Our model uses a first order KF with the assumption of constant velocity and has several hand-tuned parameters to deal with detector noise.
A threshold $\sigma_{i}=0.3$ on the detection scores as well as a minimum number of consecutive detections $d_i=3$ are necessary to start new tracks.
To keep tracks updated, another threshold $\sigma_{c}=0<\sigma_{i}$ is used, and a track can be missed for $d_m=5$ frames, before it is deleted.
Here, we choose $d_i>1$ and $d_m$ rather low to deal with the quite high false alarm rate of the detector and not initialize false tracks too early or keep them for too long.




Making a quick comparison of NN-KF to the (regressed) BIPCC baseline (cf.~Tab.~\ref{tab:all_results}), we can see a relative decrease of $44\%$ in the MOTA score.
Note that this comparison is biased, since NN-KF uses less information: only center points and no appearance information to filter false alarms.
Furthermore, NN-KF only operates in an online fashion, while BIPCC uses information of a past time window.
One should not put too much weight on this comparison, as it merely serves to anchor a classic DA-baseline, which is good enough to proceed.
Still, what strikes the eye is the high amount IDS (over $10k$) leading us to the next experiment of how the initial ID helps tracking.






\subsection{Baseline NN-tracker Started from GT}\label{sec:impl_baseline_gt}

One big concession we need to make in order to focus our attention on the integration of ReID into the Bayes filter formulation is the start of tracks.
Since one of our aims is to get rid of detection boxes, it leaves us with the question of when to start new tracks.
We leave this as future work but point out some possibilities in the discussion in Sec.~\ref{sec:discussion}.
Thus, for most of the experiments in this exploration, we make the simplifying assumption that we are given near-perfect initialization of tracks by using the first ground-truth bounding box of a person.
While this may sound like a gross simplification of the task at first, we will show here that this assumption alone is responsible for a small drop in MOTA score, because once lost, tracks cannot be resumed.
This is also closely related to the problem of single object tracking (SOT), where such an initilization is assumed.
In fact, already with a known data association, the MOT problem breaks down to multiple instances of SOT~\cite{OhMCMCDA}, when ignoring interactions between targets.
Again, we want to explore the potential of the fully integrated model, but to put it into perspective with common DA-approaches, we will also consider baseline experiments that assume the same perfect ground truth initilization.


We now first want to see how this premise will change our NN-KF baseline.
The only modification is that we start new tracks solely using the very first box of a person's ground-truth track.
This modification leads to a very poor performance (MOTA of $13.8$ with only $138$ mostly tracked targets).
The two main problems are the minimum amount of detections necessary to start a track, and that a track can not be recovered once lost.
Our goal, given the ground truth initilization, is to immediately start and then keep the track for as long as possible.
If we adjust the parameters according to this assumption ($n_i=1, \sigma_c=-0.3, d_m=90$), while the scores get better (MOTA of $34.2$, cf.~Table~\ref{tab:all_results}), even with a perfect initialization the tracker makes many mistakes.
Still, the number of FN improves, even though there is no chance to recover a lost track.
Actually, inspite of this difficulty, the number of MT targets roughly doubles and the number of ML targets halves, while the number of IDS decreases to only $1.6k$.
Of course, we do not re-initialize a track with a new ID like NN-KF, but together with the improvements in MT and ML this is still a remarkable result, regarding we are only tracking on center points.



\subsection{Training of the ReID Model}\label{sec:impl_reidmodel}
We take \emph{LuNet} from~\cite{HermansBeyer17Arxiv} as our ReID model and learn using the soft-margin formulation of the \emph{batch hard} triplet criterion~\cite{HermansBeyer17Arxiv}.
We remove those 582 track IDs that appear in the validation set from the 1812 available IDs, resulting in 1230 distinct persons which we use for training.
During training, we sample crops of $256\times96$ centered at a person's bounding box, randomly flipping and slightly wiggling them as a data augmentation, then downscale the final crop by a factor of 2 for performance reasons.

One important difference between our model and the original LuNet is that we use a $8\times3$ average-pooling layer after the last res-block, in the same way the original ResNet does.
The reason is that in this way, when convolutionally applying the network to the full image in order to compute an embedding map, we can use a stride of 1 for the average-pooling layer, resulting in high resolution embedding map.

Furthermore, because the DukeMTMC data seems to be more difficult, we have to perform curriculum learning by starting with small batches of only 2 persons with 4 images each, and gradually increasing the batch-size during training, ending up with 32 persons per batch with 4 images each.
Without curriculum learning, the network has a high probability of getting stuck during training.
Once the network is trained, its weights are fixed and no fine-tuning is performed during tracking.

\subsection{Re-ID as Appearance Model}\label{sec:impl_baseline_reid}
As mentioned in Section~\ref{sec:method_naive}, the straightforward way of utilizing a ReID model, is to use it as an appearance model to help the DA step.
Many existing works follow this direction~\cite{Kim_2015_ICCV,leal2016learning,yang2012online}, and we also report results on it.
The way this is achieved is by computing the embedding vector for the track's initial crop, which can subsequently be compared to the embedding vectors of detections.
We keep all other NN-KF parameters unchanged.

Completely replacing the spatial distance with the \emph{appearance distance} on the learned embedding space ("only ReID" in Tab.~\ref{tab:all_results}) is not able to keep the tracks for long enough.
This leads to an increase in FN and lowers FP, resulting in a slightly worse MOTA score of $29.6$ but does indeed improve the ID measures (IDS: $-271$; IDP: $+5.8$, IDF1: $+2.1$).

We then combine spatial and appearance distance, as is typical, following Eq.~\ref{eq:comb_dist}.
This combined DA results in a MOTA score of $43.2$ (``+ReID'' in Tab.~\ref{tab:all_results}), which continues the trend that was started by using only ReID.
Namely, all ID measures perform even better, and we reach the lowest number of 799 IDS, while also further decreasing FP.

Note that this already evaluates the performance of our ReID model, when used together with spatial distance for data association.
Again, the special challenge in contrast to BIPCC was to start with a ground truth initilization and keep the track for as long as possible, in an online fashion, with no chance of recovery once lost.




\subsection{Principled Integrated Tracker}\label{sec:impl_ours}
We finally turn to the main point of this work and run the fully integrated tracking model exactly as described in Section~\ref{sec:method_integration}.
This completely drops the data association problem, as well as the dependency on a person detector, by propagating the full posterior state with ID-specific measurements.
As in the previous section, we compute a track's initial embedding and compare it to the new frame's embedding map, resulting in a ReID measurement for this track.
We consider an incoming ReID measurement as missing (and thus don't perform the update step) if the smallest appearance distance in the map is above a threshold $N_\text{app}$.

Taking a look at ``Full'' in Tab.~\ref{tab:all_results}, the scores elude any comparison.
While the ID scores and the number of MT targets is still above the NN-KF, both the number of FP and FN increased dramatically, even leading to a negative MOTA.
We noticed that many of the FP are produced by embedding maps with small distances but high entropy.
By computing the entropy of the measurement-model and discarding measurements with high entropy, we are able to reduce FPs by quite a bit and improve MOTA score, cf.~``+entropy'' in Tab.~\ref{tab:all_results}.
This suggests that there is still a lot of room for improvement; in the following, we analyse typical failure cases of this model, which reveal multiple further research oppurtunities.

\section{Discussion and Research Oppurtunities}\label{sec:discussion}

\begin{figure}
	\centering
	\begin{subfigure}[b]{0.24\textwidth}
		\centering
		\includegraphics[height=2.3cm]{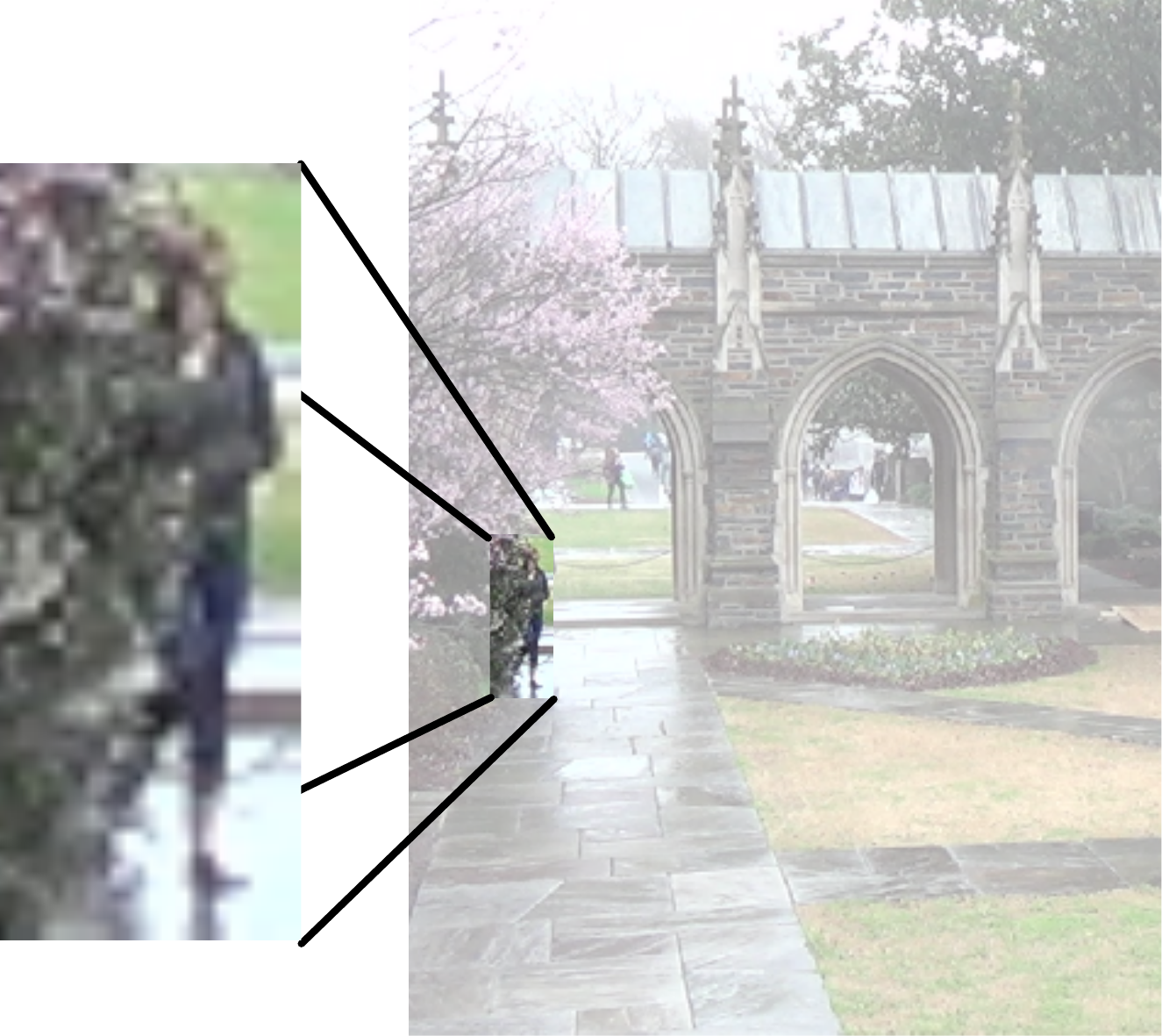}
		\caption{}
		\label{fig:failure_fn}
	\end{subfigure}%
	\begin{subfigure}[b]{0.24\textwidth}
		\centering
		\includegraphics[height=2.3cm]{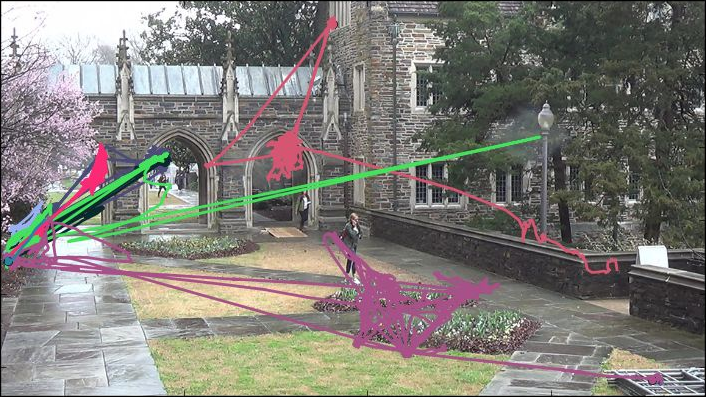}
		\caption{}
		\label{fig:failure_fp}
	\end{subfigure}
    \caption{Typical failure cases of our full integrated model: a) bad initial crops result in missed persons (FNs), b) consistent ReID mistakes in the background lead to many FPs.}
	\label{fig:failure}
\end{figure}

Our principled integration of ReID and MTMC overcomes both discrete boxes and the problem of data association and takes a further step towards end-to-end tracking.
Although it builds on the solid theoretical foundation of optimal Bayes filtering, the quantitative results still turn out to be sub-par.
Closer inspection of the results, reveal multiple major problems leaving room for a lot of improvements.
In addition, we have made several simplifying assumptions which need to be addressed in future work.

\PAR{Ground truth initilization.}
One would think that initializing tracks using ground-truth would make tracking easy.
In reality, this introduces two major difficulties: First, a strong dependence on the quality of the initial annotation.
In more than a few cases, the tracks were initialized in situations as shown in Fig.~\ref{fig:failure_fn}.
Such a start is especially bad for our fully integrated model, which relies solely on this unrepresentative initial appearance and thus ends up immediately losing the person.
Second, not only is it unrealistic to have a ground-truth initialization, but it also makes it impossible to recover a lost track.

Both problems could potentially be addressed by incorporating the notion of ``personness'' into the ReID model's training, such that it would function both as a (continuous) detector and ID-specific measurement model.
Another axis of investigation, which is already commonplace for appearance models, is to update the embedding as the track progresses, albeit avoiding drift is a topic of research of its own.

\PAR{ReID false alarms.}
Many false positives are generated after a tracked person leaves the scene, for some persons, the ReID model then starts confusing it with very specific background spots, keeping the track alive way longer than it should.
One particularly bad instance of this is shown in Fig.~\ref{fig:failure_fp}.
In theory, location-gating is taken care of by the prior, which should put zero probability mass away from the expected location.
In practice, the drift still happens due to imperfect velocity measurements and very strong FPs.

One way of reducing such false positives on the background is to incorporate background patches into the ReID model's training process.
Indeed, the ReID model used here has never seen a background during training, and it is impressive that it can still be used as such.

Another way of improving the gating from the prior is to investigate better velocity models and measurements.


\PAR{Further research oppurtunities.}
While the full integrated model gives a much richer state representation than, \eg, bounding boxes, the discretization that is necessary to evaluate it in a traditional way might incorrectly penalize full posterior maps.
Here, investigating new evaluation measures that reflect richer posteriors might capture the potential better.

Another opportunity that we have left aside in this initial research is the actual reidentification of targets across cameras.
Once the dependency of ground-truth initilization is dropped, the way opens to very interesting cross-camera continuation of tracks.

Still, there is a lot to be done before being able to learn to track from raw pixels.
We have taken one further step into this direction by clearing the path from the image into the core of the Bayesian filter algorithm through a ReID model.
Combining this work with steps towards end-to-end learning recently taken by others, inside the framework of an optimal Bayes filter, seems like a promising way forward to fully-learned tracking.







\section{Conclusion}\label{sec:conclusion}
In this work, we have presented a principled integration of person ReID and MTMC within the theoretical framework of tracking as optimal Bayes filter.
This allowed us to take one more step towards end-to-end tracking by avoiding both the problem of data association, since ID-specific measurements are created, as well as the dependency on bounding boxes, by keeping full probability maps free of any assumption about their underlying distribution.

We have shown first, yet uncompetitive results of a fully integrated method and compared them to classical DA-based baselines, some using the same strong ReID model.

This paper is of a more explorative nature and ends up asking more questions than it answers, leading way to further research opportunities.
New evaluation measures should be discussed to be able to evaluate on the full probability space, which is a much richer representation than, e.g., discrete boxes.
A general personness needs to be learned in order to start new tracks, as well as possibly height, velocity or multi-scale appearances.
First results in these directions already look promising and we plan to further investigate these issues in the future.

{\small
\bibliographystyle{ieee}
\bibliography{egbib}
}

\end{document}